\documentclass[sigconf,natbib=true]{acmart}
\usepackage{multirow}
\usepackage{balance}
\usepackage{tikz}
\usepackage{url}
\usepackage{makecell}

\def\showcomments{}
\ifdef{\showcomments}
{

}

\AtBeginDocument{%
  \providecommand\BibTeX{{%
    \normalfont B\kern-0.5em{\scshape i\kern-0.25em b}\kern-0.8em\TeX}}}

\setcopyright{acmcopyright}
\copyrightyear{2022}
\acmYear{2022}
\acmDOI{XXXXXXX.XXXXXXX}

\begin{document}

\title{Learning to Diversify for Product Question Generation}

\author{Haggai Roitman, Yotam Eshel, Alexander Nus, Eliyahu Kiperwasser}
\email{{hroitman, yeshel, anus, ekiperwasser}@ebay.com}
\affiliation{%
  \institution{eBay}
  \streetaddress{P.O. Box 1212}
  \city{Netanya}
  \country{Israel}
}

\author{Uriel Singer}
\authornote{Work done while being an intern at eBay.}
\email{urielsinger@gmail.com }
\affiliation{%
  \institution{Technion}
  \city{Haifa}
  \country{Israel}
}


\renewcommand{\shortauthors}{Roitman et al.}

\begin{abstract}
  We address the product question generation task. For a given product description, our goal is to generate questions that reflect potential user information needs that are either missing or not well covered in the description. Moreover, we wish to cover diverse user information needs that may span a multitude of product types. To this end, we first show how the T5 pre-trained Transformer encoder-decoder model can be fine-tuned for the task.  Yet, while the T5 generated questions have a reasonable quality compared to the state-of-the-art method for the task (KPCNet), many of such questions are still too general, resulting in a sub-optimal global question diversity. As an alternative, we propose a novel learning-to-diversify (LTD) fine-tuning approach that allows to enrich the language learned by the underlying Transformer model. Our empirical evaluation shows that, using our approach significantly improves the global diversity of the underlying Transformer model, while preserves, as much as possible, its generation relevance. 
\end{abstract}

\ccsdesc[500]{Computing methodologies~Natural language generation}
\ccsdesc[500]{Applied computing~Electronic commerce}

\keywords{product, question generation, neural networks, diversification}


\maketitle

\section{Introduction}\label{sec:intro}
\begin{table}[t]
\caption{Motivating example: questions generated by three different generative models for two products from the \emph{Home \& Kitchen} category}
\center\setlength\tabcolsep{3.0pt}
\small
\begin{tabular}{|c|l|}
\hline
\textbf{Product} & \multicolumn{1}{l|}{\emph{tuft \& needle five handcrafted mattress ( twin )}}                                                                                                                                                     \\ \hline
KPCNet           & \multicolumn{1}{l|}{\begin{tabular}[c]{@{}l@{}}will this mattress fit a mattress ?\\ will this mattress fit a queen mattress ? 
\end{tabular}}                             
\\ \hline
T5               & \multicolumn{1}{l|}{\begin{tabular}[c]{@{}l@{}}what is the warranty on this mattress ? \\ what are the dimensions of this item ? 
\end{tabular}}                                             \\ \hline
T5+LTD             & \multicolumn{1}{l|}{\begin{tabular}[c]{@{}l@{}}what is the warranty on this mattress ? 
\\ does it come with a cover to protect the \\mattress from spills ?\end{tabular}}                     \\ \hline\hline

\textbf{Product} & \multicolumn{1}{l|}{\begin{tabular}[c]{@{}l@{}}\emph{gibson couture bands 16-piece dinnerware set, blue and cream}\end{tabular}}                                                                          \\ \hline
KPCNet           & \multicolumn{1}{l|}{\begin{tabular}[c]{@{}l@{}}are these plates made in the usa ? \\ what are the dimensions of the set ?
\end{tabular}}        
\\ \hline
T5               & \multicolumn{1}{l|}{\begin{tabular}[c]{@{}l@{}}what is the diameter of the dinner plate ? \\ what is the size of the bowls ?
\end{tabular}}                   \\ \hline
T5+LTD             & \multicolumn{1}{l|}{\begin{tabular}[c]{@{}l@{}}are they dishwasher safe ? \\ where is this product made ?
\end{tabular}}                                                                                \\ \hline
\end{tabular}\label{fig:motivation}
\end{table}

E-Commerce is fast-expanding, with a never-ending requirement to offer personalized shopping experiences to users. Product descriptions on E-Commerce websites, such as Amazon, eBay and Shopify, serve as an important knowledge source to potential buyers for making purchase decisions. Product descriptions strive to be as informative and accurate as possible, trying to satisfy a variety of user information needs. In reality, creating product descriptions that can satisfy any possible information need is extremely difficult, as it is hard to anticipate in advance the full range of such needs. 
The gap between a buyer's information need and the information available in a product description, usually requires the buyer to contact the seller directly with clarification questions or to forfeit her purchase intent, which leads to an undesirable churn.




In this work, we aim to mitigate such gaps by automatically generating product clarification questions to be recommended to sellers. Such a recommendation can take place, for example, already during the product's listing process once the seller provides the product's description. As a result, the seller may revise the product description with missing details. 
Automatically generated questions should be as \textbf{relevant} and \textbf{diverse} as possible, covering a multitude of informative aspects that are specific to the product and its usage. 

Generating questions that are both relevant and diverse with respect to a specific product is a challenging task. This becomes even more challenging when facing 
a wide range of information needs over a multitude of product types~\cite{Zhang-KPCNet}. As an illustrative example, Table~\ref{fig:motivation} shows two products (only product titles and their images are provided for brevity) and the top-2 questions that were generated by three state-of-the-art generative models: KPCNet~\cite{Zhang-KPCNet}, T5~\cite{colin-2020-T5} and T5+LTD -- our proposed solution. Ignoring 
obvious mistakes such as illogical questions (e.g., \emph{will this mattress fit a mattress ?}), we can observe that, by trying to cover as many product types as possible, some models may ``prefer'' to generate questions that are too general, yet still relevant in a way (e.g., \emph{what are the dimensions of <product>?}). As this example demonstrates, one may wish to expand the range of questions that can be generated over a given products collection (\emph{global diversity}), while still preserving the ability to generate questions that are both relevant and diverse to a specific product in the collection (\emph{local diversity}).

Trying to address the challenge, in this work, we propose a novel \emph{learning-to-diversify} (LTD) fine-tuning approach for product question generation using Transformers~\cite{vaswani2017attention}. 
To this end, using a bi-branch network architecture, we fine-tune the underlying pre-trained T5 Transformer encoder-decoder model by training it in a pairwise-way with multiple question-pairs per product. Our goal is to maximize the generation likelihood of each pair of questions, while at the same time, minimize their semantic similarity. The semantic similarity between a pair of questions is measured with respect to the latent query representations learned by the Transformer's decoder. Applying our approach on the underlying Transformer model requires no further change during inference time. Utilizing such a learning approach allows to enhance the underlying Transformer model's ability to generate diverse questions which cover a much broader range of product information needs, resulting in an increase in its global (question generation) diversity. This is done, while still preserving (as much as possible) the underlying Transformer model's ability to generate diverse questions that are relevant to a specific product in the collection, hence, preserving its local (question generation) diversity. 

Using product descriptions and questions from different product categories on Amazon, we demonstrate that, the quality of questions that are generated using our fine-tuning approach (LTD) is better than of those generated by several alternative models, including the underlying pre-trained T5 Transformer model when it is fine-tuned in the ``traditional'' way. 

The rest of this paper is organized as follows. We discuss related works in Section~\ref{sec:related} and present our learning framework in Section~\ref{sec:model}. We report our evaluation in Section~\ref{sec:eval} and conclude in Section~\ref{sec:summary}.

\section{Related Work}\label{sec:related}
We review works primarily related to
either product or diversified question generation tasks. 
A more general overview on the question generation task in NLP can be found in~\cite{QGNLP2021}. 

\subsection{Product question generation}
The product question generation task is a relatively new task. 
A common approach, is to model the generation process as a sequence-to-sequence (seq2seq) setting, using the product's description as the source text to be encoded and the required question as the target text to be decoded~\cite{QGNLP2021}. Yet, vanilla recurrent-neural networks that are applied to the task suffer from common problems such as unknown words and difficulty to control the specificity and diversity of generated questions~\cite{li-etal-2016-diversity}. 
Xiao at el.~\cite{xiao2019question} have handled unknown words (yet not diversity) using a pointer-generator network. Zhang et al.~\cite{Zhang-KPCNet} have proposed KPCNet -- a seq2seq model that attended on selected product description keywords to improve question specificity. 
Several other works~\cite{rao-daume-iii-2019-answer,yu-etal-2020-review,wang2021harvest} have utilized adversarial learning (e.g., GANs) to improve question generation ``quality'' by using an additional discriminator model. Yet, training such a discriminator requires additional labeled data with question answers. 
Wang et al.~\cite{wang2021harvest} have further suggested to train the discriminator with question pairs, consistenting of a true question and a negatively sampled one. Yet, the generator in~\cite{wang2021harvest} heavily depends on the availability of auxiliary data such as product properties and user interest aspects. Finally, Majumder et al.~\cite{majumder2021ask} have utilized global product knowledge to predict missing aspects. 

\subsection{Diversified question generation}
Enhancing the diversity of text generation, and questions in particular, was the aim of several previous works~\cite{li-etal-2016-diversity,ippolito-etal-2019-comparison,shen2019mixture,cao-etal-2019-controlling, cho-etal-2021-contrastive,ShaoApex2021}. 
A common approach to diversify the generated questions is to apply diversification methods during model inference~\cite{ippolito-etal-2019-comparison}. Common methods include: diverse beam-search (DBS), top-p and/or top-k sampling and post-generation analysis (e.g., clustering~\cite{Zhang-KPCNet}, specificity classification~\cite{cao-etal-2019-controlling}). 
Yet, such inference methods strongly 
depend on the language learning capacity of the underlying trained model. 
An alternative approach is, therefore, to allow the model to ``explore'' more during its training phase~\cite{li-etal-2016-diversity}. To the best of our knowledge, only few related works have focused on such an approach. 
Shen et al.~\cite{shen2019mixture} have trained a mixture of experts model to learn different generation styles. Shao et al.~\cite{ShaoApex2021} have proposed Apex -- a Conditional Variational Auto-encoder for product description generation from few keywords. The trade-off between accuracy and diversity  was controlled by setting a bound on the KL-divergence loss. 
Cho et al.~\cite{cho-etal-2021-contrastive} have employed contrastive learning for question generation over multiple documents. To this end, their model was trained on triplets containing a single training question with a pair of positive and negative document sets. The generator's goal was to generate questions that are only grounded in the positive documents. 

\subsection{Main differences}
The main goal of our work is to improve the global diversity of questions generated by an underlying pre-trained Transformer~\cite{vaswani2017attention} model. Most existing works have fine-tuned pre-trained Transformer models with the primary objective of maximizing question generation likelihood; while diversity was commonly dealt only as a secondary objective during inference time~\cite{QGNLP2021,li2022survey}.  Compared to~\cite{majumder2021ask}, which have utilized the Transformer model for the same task, we do not use any auxiliary data. Finally, our method is eminently different from existing contrastive learning methods~\cite{cho-etal-2021-contrastive}, as in our case there are only positive examples. To the best of our knowledge, we are not aware of any similar work that has fine-tuned pre-trained Transformer models for enhanced diversity as we do.


\section{Framework}\label{sec:model}
We first formally define the 
product question generation (PQG) task (Section~\ref{sec:task}).
We then shortly discuss how the T5 pre-trained Transformer encoder-decoder model~\cite{colin-2020-T5} can be fine-tuned for this task (Section~\ref{sec:transfromers}).
Finally, we introduce our alternative learning-to-diversify (LTD) fine-tuning approach (Section~\ref{sec:method}). 

\subsection{Product Question Generation Task}\label{sec:task}
For a given product description text sequence, termed hereinafter as ``context'', $c=(x_1,x_2,\ldots,x_n)$, the goal of the PQG task is to generate a question text sequence $q=(y_1,y_2,\ldots,y_m,?)$. The generated question $q$ should be informative enough to clarify details related to the product (e.g., an inquiry about a specific product aspect such as its color or size, usage, compatibility, etc) which are not (fully) described in $c$. Commonly, several questions $Q_c=\{q_1,q_2,\ldots,q_k\}$ may be generated for a given context $c$. The questions-set $Q_c$ should be as diverse as possible, covering different potential user information needs regarding the product. In that case, we say that $Q_c$ is \emph{locally diverse}. For a given products collection $C=\{c_1,c_2,\ldots,c_g\}$, we further wish that the corresponding generated questions super-set $Q_{C}=\bigcup_{c\in C}{Q_c}$ would be as diverse as possible, covering a wide range of information needs over all products in the collection $C$. In that case, we say that $Q_{C}$ is \emph{globally diverse}. 

\subsection{Fine-tuning the T5 model for the PQG task}\label{sec:transfromers}
In this work, we utilize the T5~\cite{colin-2020-T5} pre-trained Transformer~\cite{vaswani2017attention} encoder-decoder model and fine-tune it for the PQG task.
``Traditionally'', model fine-tuning is implemented as a \emph{conditional generation} (seq2seq) task~\cite{QGNLP2021}. Formally, for a given training sample $(c,q)$, the goal is to maximize the generation likelihood:
\vspace{-0.08in}
\begin{equation}
    p(q|c,\theta)=\prod_{t=1}^{m}p(y_t|q_{<t},c,\theta),
\end{equation}
where $q_{<t}$ denotes the question words that were generated up to step $t$ and $\theta$ represents the Transformer model parameters. 

Next, we shortly describe how the Transformer encoder-decoder model can be utilized for this task. Before we move on, it is important to mention that, several Transformer ``block'' layers can be stacked together to increase the model's learnability~\cite{vaswani2017attention}. For full technical details on the Transformer model, we kindly refer the reader to~\cite{vaswani2017attention}. Given context $c$, the Transformer encoder first encodes it into a latent representation $h_e=Trans_{enc}(c)$. At step $t$, the Transformer decoder block at layer $l\in\{1,2,\ldots,L\}$ attends both on $h_e$ and the output of the previous layer $h_d^{(l-1)}$, and then outputs a representation $h_d^{(l)}=Trans_{dec}(h_e,h_d^{(l-1)})$; with $h_d^{(0)}=q_{<t}$. The next query word $y_t$ is then generated using a conditional generation ``head'' (CG-head\footnote{Implemented using a feed-forward (FF) layer, followed by a softmax operation~\cite{vaswani2017attention}.}), which transforms $h_d^{(L)}$ into a distribution over the Transformer's vocabulary and assigns $y_t$ as the word with the highest likelihood. Training is usually implemented using a \emph{teacher-forcing} approach~\cite{QGNLP2021}, where at each step $t$, the training question words up to step $t$ are used instead of $q_{<t}$. To simplify notation, from now on, we shall abbreviate $h_d$ as $h$.

\begin{figure}[t!]
\input{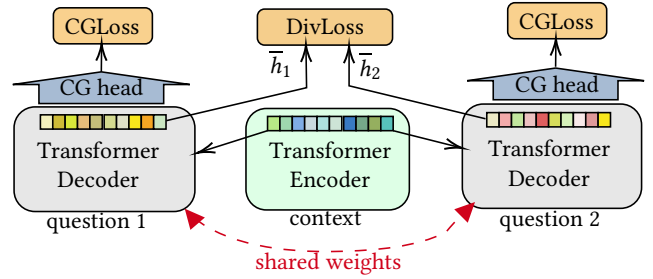}
\caption{Learning-to-diversify for PQG task}\label{fig:model}
\end{figure}

\subsection{Learning to Diversify}\label{sec:method}
We next propose an alternative (fine-tuning) approach whose primary goal is to enhance the overall (global) diversity of questions generated by the underlying Transformer model.

Our approach is built on the hypothesis, which is empirically verified later on using the T5 model, that, pre-trained Transformer models for the PQG task may tend to learn common (general) questions that appear in the training set in the expense of more rare ones. Therefore,
we wish to improve the “exploration” capability of the underlying Transformer model by allowing it
to learn a more flexible language model that results in a generation of more (globally) diverse questions.

Our proposed learning-to-diversify (LTD) fine-tuning approach is based on a \emph{bi-branch} network architecture and is illustrated in  Figure~\ref{fig:model} (assuming $L=1$ for simplicity) and works as follows. We train the underlying Transformer model on triplets $(c,q_1,q_2)$, where $q_1$ and $q_2$ are a pair of different target questions that the model needs to generate for a given input context $c$. This part is simply implemented as before, where for each question $q_i$ ($i\in\{1,2\}$), we generate the next word $y_t^{i}$ based on the Transformer decoder representation $h_i$, respectively. Let $CGLoss^{i}_t$ be the overall corresponding conditional generation (CG) loss\footnote{Implemented as cross-entropy loss~\cite{QGNLP2021}.} 
incurred by generating the question words $\hat{q}_i=(y^{i}_{1},\ldots,y^{i}_{m_{i}})$.

Next, we ``infuse'' exploration to the underlying model by encouraging it to generate two questions that are eminently semantically different from each other. To this end, following the \textbf{teacher-forcing} approach, for each question $q_i$ we set $h_i^{(0)}=q_i$ and obtain its representations $h_i^{(l)}$ over all the Transformer decoder layers. Here we note that, the key idea behind such an approach, is to obtain a given question's representation by the decoder assuming that the model \textbf{has correctly generated it}. Our goal is therefore, to allow a backward feedback to the model based on how different are the two questions representations are. 
The overall difference between the two questions representations by the decoder is measured according to the cosine loss term between the representations that were obtained by the Transformer decoder with $L$ blocks (stack):
$DivLoss^{1,2}=\frac{1}{L}\sum_{l=1}^{L}cosine($ ${\bar{h}}_{1}^{(l)},{\bar{h}}_{2}^{(l)})$, where ${\bar{h}}_{i}^{(l)}$ is calculated using \emph{mean-pooling} over the sequence dimension\footnote{$h_{i}^{(l)}$ is a matrix defined over the input (sequence) and embedding dimensions~\cite{vaswani2017attention}.} of  $h_{i}^{(l)};i\in\{1,2\}$, respectively. 

Finally, we use the diversity loss as a regularization term for the two CG losses, as follows: 
\vspace{-0.05in}
\begin{equation}
    Loss = CGLoss^{1} + CGLoss^{2} + \lambda\cdot DivLoss^{1,2},
\end{equation}

where the hyperparameter $\lambda>0$ controls to what extent we wish the model to explore towards diversification.

\section{Evaluation}\label{sec:eval}
\subsection{Experimental Setup}
\subsubsection{Datasets}

\begin{table}[t!]
\caption{Datasets used for the evaluation}
\center\setlength\tabcolsep{2.0pt}
\small
\begin{tabular}{|l|c|c|c|c|c|}
\hline
\multicolumn{1}{|c|}{\textbf{Category}} & \textbf{\#Products} & \textbf{\#Questions} & \textbf{Train} & \multicolumn{1}{l|}{\textbf{Validation}} & \textbf{Test} \\ \hline
\emph{Home \& Kitchen}                         & 23,859              &  145,536                    & 19,119         & 2,435                                    & 2,305         \\ \hline
\emph{Office Products}                         & 2,731               &   13,775                   & 2,190          & 285                                      & 256           \\ \hline
\emph{Sports \& Outdoors}                      & 8,398               &  54,383                    & 6,664          & 834                                      & 835           \\ \hline
\emph{Electronics}                             & 23,900              &     166,182                 & 19,108         & 2,389                                    & 2,389         \\ \hline
\end{tabular}\label{tab:datasets}
\end{table}

We summarize the details of the datasets that we use for our evaluation in Table~\ref{tab:datasets}. All datasets are based on products sold on Amazon and questions that were asked by Amazon buyers~\cite{amazondata}.  Following previous works~\cite{Zhang-KPCNet,rao-daume-iii-2019-answer}, we use product-question pairs sampled from the \emph{Home \& Kitchen} and the \emph{Office Products} categories of the Amazon dataset. We use the pre-processed dataset version of~\cite{Zhang-KPCNet} for both categories. We further extend our evaluation with product-question pairs that are sampled from two additional categories, namely: \emph{Sports \& Outdoors} and \emph{Electronics}. Following~\cite{Zhang-KPCNet}, each product context consists of the concatenation of the product title and description. On average, on each dataset, each product context has about 3-10 questions. On each dataset, we use about 80\% of the products for training, 10\% for validation (tuning) and the last 10\% for testing. 

\subsubsection{Baselines}
Our first line of baselines are those that were previously evaluated in~\cite{Zhang-KPCNet}. This includes: \textbf{MLE} -- a vanilla seq2seq model~\cite{Zhang-KPCNet}, \textbf{hMup}~\cite{shen2019mixture} -- a mixture of experts model, and \textbf{KPCNet}~\cite{Zhang-KPCNet} -- the current state-of-the-art method for the PQG task. For a fair comparison, we use the best baseline results\footnote{Baselines code and results are available in: \url{https://github.com/blmoistawinde/KPCNet}} that are provided by~\cite{Zhang-KPCNet}.  

Since we apply our learning-to-diversify (LTD) fine-tuning approach on the pre-trained T5~\cite{colin-2020-T5} model, we further evaluate the same pre-trained model when it is fine-tuned using the ``traditional'' approach  (Section~\ref{sec:transfromers}), denoted \textbf{T5}. Finally, we evaluate \textbf{T5+LTD}, the same T5 pre-trained model fine-tuned with our LTD approach. 

\subsubsection{Implementation and Training}
We implement both fine-tuning approaches
with pytorch and huggingface\footnote{\url{https://huggingface.co/docs/transformers/index}} pre-trained T5 model (``t5-base''). 
We use the Adam~\cite{kingma2015adam} optimizer, with a learning rate of $10^{-4}$, a batch size of $8$, trained for $3$ epochs (saving the best checkpoint using the validation-set) with a machine with $4$ GPUs. For a fair comparison, we fix the inference of all Transformer-based models (i.e., T5 and T5+LTD) and use the diversity beam-search (DBS)~\cite{Vijayakumar2016DiverseBS} method; which gives the best inference quality (on the validation-set) compared to other alternative inference methods~\cite{ippolito-etal-2019-comparison}. 
We tune (on the validation-set) $\lambda\in(0,1]$, with $\lambda=0.1$ derived as the ``best'' hyperparmeter value choice over all categories. Following~\cite{Zhang-KPCNet}, using the validation-set, we set the number of beam groups to $3$ (with $6$ total questions generated per product), length penalty of $1.0$, diversity penalty of $5.0$ and non-repeat of bi-grams.

\subsubsection{Metrics}
We evaluate both the \textbf{relevance} and the \textbf{diversity} of the generated questions. Following~\cite{Zhang-KPCNet}, for each product, we evaluate the top-$3$ generated questions. We measure the generated questions relevance along two main dimensions, namely: \textbf{lexical} (based on surface-level lexical overlap) and \textbf{semantic} (based on word-context). Following~\cite{Zhang-KPCNet}, we measure lexical relevance using the (top-$1$ question) \textbf{BLEU}~\cite{bleu}, (top-$3$ questions) \textbf{Avg-BLEU}~\cite{Zhang-KPCNet} and (top-$1$ question) \textbf{METEOR}~\cite{banerjee-lavie-2005-meteor} metrics. Following~\cite{bert-score}, we measure semantic relevance using the \textbf{BERTScore}\footnote{\url{https://github.com/Tiiiger/bert_score}} (top-$1$ question) and \textbf{Avg-BERTScore} (top-$3$ questions) metrics. 

We measure the diversity of generated questions both \textbf{locally} (per-single product) and \textbf{globally} (per products-collection). Similar to the relevance metrics, we use both lexical and semantic measures. Following~\cite{Zhang-KPCNet}, we measure the lexical local diversity according to Pairwise-BLEU (abbreviated as \textbf{PW-BLEU} in our tables) (top-$3$ questions). This is an extension of the Self-BLEU metric~\cite{shu-etal-2019-generating}, having every time one question (out of top-$3$) being considered as the ``hypothesis'' and the rest as ``references''~\cite{Zhang-KPCNet}. As a semantic alternative of this measure, we further calculate Pairwise-BERTScore (abbreviated as \textbf{PW-BERTScore} in our tables), where we replace BLEU with BERTScore. Here we note that, since the goal is to have three questions per product that are different from each other, \textbf{lower} PW-BLEU and PW-BERTScore values translate to better local diversity.

We further measure lexical global diversity of generated questions using the Distinct-N~\cite{li-etal-2016-diversity} metric (abbreviated as \textbf{Dist-N} in our tables); where $N\in\{1,2,3\}$ denotes the N-gram size. This metric is calculated by counting (considering all top-$1$ questions generated for the test-set products) the number of unique N-grams, divided by the total number of N-grams~\cite{li-etal-2016-diversity}. Since this metric is lexical in its nature (i.e., counts exact words), we further wish to evaluate global diversity using a more semantic measure. Following~\cite{lai-etal-2020-diversity}, we use the embedding-based diversity measure (abbreviated as \textbf{e-Div} in our tables). To this end, given a collection of question embeddings, for each embedding dimension, we first calculate its radius (the standard-deviation of values in that dimension~\cite{lai-etal-2020-diversity}). The embedding-based diversity (e-Div) is then calculated as the geometric mean of the radius values. Hence, the larger is the radius over each dimension, the more spread is the questions embedding space, and therefore, more (globally) diverse. Finally, we obtain the question embeddings (with 512 dimensions) using a SentenceTransformer~\cite{reimers2019sentencebert} encoder dedicated for paraphrasing tasks. 


\begin{table}[t!]
\caption{Comparison of question generation relevance obtained by the various baselines (\emph{Home \& Kitchen category})}
\center\setlength\tabcolsep{2.0pt}
\small
\begin{tabular}{|l|ccc||cc|}
\hline
       & \multicolumn{3}{c||}{\textbf{Lexical Relevance}}                                       & \multicolumn{2}{c|}{\textbf{Semantic Relevance}}     \\ \hline
       & \multicolumn{1}{c|}{BLEU$\uparrow$}          & \multicolumn{1}{c|}{Avg-BLEU$\uparrow$}      & \multicolumn{1}{c||}{METEOR$\uparrow$}        & \multicolumn{1}{c|}{BERTScore$\uparrow$}      & Avg-BERTScore$\uparrow$  \\ \hline
MLE    & \multicolumn{1}{c|}{18.1}          & \multicolumn{1}{c|}{16.9}          & 14.9          & \multicolumn{1}{c|}{30.9}          & 30.9          \\ \hline
hMup   & \multicolumn{1}{c|}{17.8}          & \multicolumn{1}{c|}{9.9}           & 15.4          & \multicolumn{1}{c|}{23.5}          & 28.4          \\ \hline
KPCNet & \multicolumn{1}{c|}{17.8}          & \multicolumn{1}{c|}{16.2}          & 16.2          & \multicolumn{1}{c|}{32.3}          & 31.6          \\ \hline\hline
T5     & \multicolumn{1}{c|}{19.2}          & \multicolumn{1}{c|}{17.1}          & \textbf{16.4} & \multicolumn{1}{c|}{32.1}          & 31.6          \\ \hline
T5+LTD    & \multicolumn{1}{c|}{\textbf{20.1}} & \multicolumn{1}{c|}{\textbf{17.2}} & 16.3          & \multicolumn{1}{c|}{\textbf{32.9}} & \textbf{31.8} \\ \hline
\end{tabular}\label{tab:relevance all}
\end{table}

\begin{table}[t!]
\caption{Comparison of question generation diversity obtained by the various baselines (\emph{Home \& Kitchen category})}
\center\setlength\tabcolsep{2.0pt}
\small
\begin{tabular}{|l|cc||cccc|}
\hline
       & \multicolumn{2}{c||}{\textbf{Local Diversity}}      & \multicolumn{4}{c|}{\textbf{Global Diversity}}                                                                                              \\ \hline
       & \multicolumn{1}{c|}{PW-BLEU$\downarrow$}       & \multicolumn{1}{c||}{PW-BERTScore $\downarrow$}  & \multicolumn{1}{c|}{Dist-1$\uparrow$}           & \multicolumn{1}{c|}{Dist-2$\uparrow$}            & \multicolumn{1}{c|}{Dist-3$\uparrow$}            & \multicolumn{1}{l|}{e-Div$\uparrow$} \\ \hline
MLE    & \multicolumn{1}{c|}{26.8}          & 61.6          & \multicolumn{1}{c|}{2.7}          & \multicolumn{1}{c|}{5.7}           & \multicolumn{1}{c|}{7.8}           & 29.7                       \\ \hline
hMup   & \multicolumn{1}{c|}{\textbf{13.4}} & \textbf{44.5} & \multicolumn{1}{c|}{2.3}          & \multicolumn{1}{c|}{9.6}            & \multicolumn{1}{c|}{11.1}          & 26.8                       \\ \hline
KPCNet & \multicolumn{1}{c|}{42.8}          & 75.2          & \multicolumn{1}{c|}{3.4}          & \multicolumn{1}{c|}{9.6}           & \multicolumn{1}{c|}{15.3}          & 29.9                       \\ \hline\hline
T5     & \multicolumn{1}{c|}{27.1}          & 58.2          & \multicolumn{1}{c|}{4.1}          & \multicolumn{1}{c|}{9.2}           & \multicolumn{1}{c|}{14.3}          & 29.8                       \\ \hline
T5+LTD    & \multicolumn{1}{c|}{26.2}          & 54.0          & \multicolumn{1}{c|}{\textbf{5.1}} & \multicolumn{1}{c|}{\textbf{11.5}} & \multicolumn{1}{c|}{\textbf{17.5}} & \textbf{30.4}              \\ \hline
\end{tabular}\label{tab:diversity all}
\end{table}

\begin{table*}[tbh]\caption{Comparison of question generation quality between T5 and T5+LTD for different product categories. The percentages bellow the T5+LTD metric values denote the relative improvement/degradation compared to T5.} 
\center\setlength\tabcolsep{2.0pt}
\small
\begin{tabular}{|ll|ccc|cc||cc|cccc|}
\hline
\multicolumn{2}{|c|}{\multirow{2}{*}{\textbf{Category}}}                  & \multicolumn{3}{c|}{\textbf{Lexical Relevance}}                                       & \multicolumn{2}{c||}{\textbf{Semantic Relevance}}   & \multicolumn{2}{c|}{\textbf{Local Diversity}}      & \multicolumn{4}{c|}{\textbf{Global Diversity}}                                                                               \\ \cline{3-13} 
\multicolumn{2}{|c|}{}                                                    & \multicolumn{1}{c|}{BLEU$\uparrow$}          & \multicolumn{1}{c|}{Avg-BLEU$\uparrow$}      & METEOR $\uparrow$       & \multicolumn{1}{c|}{BERTScore$\uparrow$}     & Avg-BERTScore$\uparrow$ & \multicolumn{1}{c|}{PW-BLEU$\downarrow$}       & PW-BERTScore$\downarrow$   & \multicolumn{1}{c|}{Dist-1$\uparrow$}        & \multicolumn{1}{c|}{Dist-2$\uparrow$}        & \multicolumn{1}{c|}{Dist-3$\uparrow$}        & e-Div$\uparrow$         \\ \hline
\multicolumn{1}{|l|}{\multirow{2}{*}{\textit{Home \& Kitchen}}}    & T5     & \multicolumn{1}{c|}{19.2}          & \multicolumn{1}{c|}{17.1}          & \textbf{16.4} & \multicolumn{1}{c|}{32.1}          & 31.6          & \multicolumn{1}{c|}{27.1}          & 58.2          & \multicolumn{1}{c|}{4.1}           & \multicolumn{1}{c|}{9.2}           & \multicolumn{1}{c|}{14.3}          & 29.8          \\ \cline{2-13} 
\multicolumn{1}{|c|}{}                                           & T5+LTD & \multicolumn{1}{c|}{\makecell{\textbf{20.1}\\\scriptsize{(+4.7\%)}}} & \multicolumn{1}{c|}{\makecell{\textbf{17.2}\\\scriptsize{(+0.6\%)}}} & \makecell{16.3\\\scriptsize{(-0.6\%)}}          & \multicolumn{1}{c|}{\makecell{\textbf{32.9}\\\scriptsize{(+2.5\%)}}} & \makecell{\textbf{31.8}\\\scriptsize{(+0.6\%)}} & \multicolumn{1}{c|}{\makecell{\textbf{26.2}\\\scriptsize{(+3.4\%)}}} & \makecell{\textbf{54.0}\\\scriptsize{(+7.8\%)}} & \multicolumn{1}{c|}{\makecell{\textbf{5.1}\\\scriptsize{(+24.4\%)}}}  & \multicolumn{1}{c|}{\makecell{\textbf{11.5}\\\scriptsize{(+25.0\%)}}} & \multicolumn{1}{c|}{\makecell{\textbf{17.5}\\\scriptsize{(+22.4\%)}}} & \makecell{\textbf{30.4}\\\scriptsize{(+2.0\%)}} \\ \hline\hline
\multicolumn{1}{|l|}{\multirow{2}{*}{\textit{Office Products}}}  & T5     & \multicolumn{1}{c|}{\textbf{16.9}}          & \multicolumn{1}{c|}{\textbf{13.6}} & 14.9          & \multicolumn{1}{c|}{\textbf{32.2}} & \textbf{30.9} & \multicolumn{1}{c|}{28.2}          & 69.6          & \multicolumn{1}{c|}{12.2}          & \multicolumn{1}{c|}{21.7}          & \multicolumn{1}{c|}{27.9}          & 25.2          \\ \cline{2-13} 
\multicolumn{1}{|c|}{}                                           & T5+LTD & \multicolumn{1}{c|}{\textbf{16.9}}          & \multicolumn{1}{c|}{\makecell{13.5\\\scriptsize{(-0.7\%)}}}          & \makecell{\textbf{15.3}\\\scriptsize{(+2.7\%)}} & \multicolumn{1}{c|}{\makecell{31.0\\\scriptsize{(-3.9\%)}}}          & \makecell{29.0\\\scriptsize{(-6.7\%)}}          & \multicolumn{1}{c|}{\makecell{\textbf{27.2}\\\scriptsize{(+3.7\%)}}} & \makecell{\textbf{64.9}\\\scriptsize{(+7.2\%)}} & \multicolumn{1}{c|}{\makecell{\textbf{14.5}\\\scriptsize{(+18.8\%)}}} & \multicolumn{1}{c|}{\makecell{\textbf{28.1}\\\scriptsize{(+29.5\%)}}} & \multicolumn{1}{c|}{\makecell{\textbf{35.6}\\\scriptsize{(+41.3\%)}}} & \makecell{\textbf{25.8}\\\scriptsize{(+2.4\%)}} \\ \hline\hline
\multicolumn{1}{|l|}{\multirow{2}{*}{\textit{Sports \& Outdoors}}} & T5     & \multicolumn{1}{c|}{\textbf{4.5}}  & \multicolumn{1}{c|}{2.5}           & \textbf{10.2} & \multicolumn{1}{c|}{\textbf{23.1}} & \textbf{22.5} & \multicolumn{1}{c|}{\textbf{32.5}} & 69.2          & \multicolumn{1}{c|}{7.1}           & \multicolumn{1}{c|}{13.9}          & \multicolumn{1}{c|}{18.8}          & 25.2          \\ \cline{2-13} 
\multicolumn{1}{|c|}{}                                           & T5+LTD & \multicolumn{1}{c|}{\makecell{4.4\\\scriptsize{(-2.3\%)}}}           & \multicolumn{1}{c|}{\makecell{\textbf{2.6}\\\scriptsize{(+4.0\%)}}}  & \makecell{10.1\\\scriptsize{(-1.0\%)}}          & \multicolumn{1}{c|}{\makecell{23.0\\\scriptsize{(-0.4\%)}}}          & \makecell{22.4\\\scriptsize{(-0.4\%)}}          & \multicolumn{1}{c|}{\makecell{34.2\\\scriptsize{(-5.2\%)}}}          & \makecell{\textbf{68.4}\\\scriptsize{(+1.2\%)}} & \multicolumn{1}{c|}{\makecell{\textbf{8.5}\\\scriptsize{(+19.7\%)}}}  & \multicolumn{1}{c|}{\makecell{\textbf{17.1}\\\scriptsize{(+23.0\%)}}} & \multicolumn{1}{c|}{\makecell{\textbf{23.2}\\\scriptsize{(+23.4\%)}}} & \makecell{\textbf{25.7}\\\scriptsize{(+2.0\%)}} \\ \hline\hline
\multicolumn{1}{|l|}{\multirow{2}{*}{\textit{Electronics}}}      & T5     & \multicolumn{1}{c|}{4.6}           & \multicolumn{1}{c|}{\textbf{2.8}}           & 9.6           & \multicolumn{1}{c|}{20.5}          & 20.4          & \multicolumn{1}{c|}{\textbf{39.8}} & 72.5          & \multicolumn{1}{c|}{3.6}           & \multicolumn{1}{c|}{6.9}           & \multicolumn{1}{c|}{9.5}           & 23.1          \\ \cline{2-13} 
\multicolumn{1}{|c|}{}                                           & T5+LTD & \multicolumn{1}{c|}{\makecell{\textbf{4.9}\\\scriptsize{(+6.5\%)}}}  & \multicolumn{1}{c|}{\textbf{2.8}}           & \makecell{\textbf{10.3}\\\scriptsize{(+7.3\%)}} & \multicolumn{1}{c|}{\makecell{\textbf{22.3}\\\scriptsize{(+8.8\%)}}} & \makecell{\textbf{21.4}\\\scriptsize{(+4.9\%)}} & \multicolumn{1}{c|}{\makecell{40.4\\\scriptsize{(-1.5\%)}}}          & \makecell{\textbf{71.5}\\\scriptsize{(+1.4\%)}} & \multicolumn{1}{c|}{\makecell{\textbf{3.7}\\\scriptsize{(+2.8\%)}}}  & \multicolumn{1}{c|}{\makecell{\textbf{7.9}\\\scriptsize{(+14.5\%)}}}  & \multicolumn{1}{c|}{\makecell{\textbf{11.0}\\\scriptsize{(+15.8\%)}}} & \makecell{\textbf{24.0}\\\scriptsize{(+3.9\%)}}\\ \hline
\end{tabular}\label{tab:categories}
\end{table*}

\subsection{Results}\label{sec:results}
We now summarize the results of our empirical evaluation. Our goal is to answer the following three main research questions (RQs):
\begin{itemize}
    \item \textbf{RQ1}: How good (both in terms of relevance and diversity) are the questions generated by the T5 baseline  
    compared to those generated by the best previously performing method (KPCNet) for the PQG task?
    \item \textbf{RQ2}: How good are the questions generated by 
    T5+LTD compared to T5?
    \item \textbf{RQ3}: How more globally diverse are questions generated by T5+LTD compared to T5?
\end{itemize}

\subsubsection{RQ1: ``Traditional''  pre-trained T5 model fine-tuning}
To answer RQ1, we compare all baselines over the \emph{Home \& Kitchen} category, for which the \textbf{best} generated questions of the MLE, hMup and KPCNet baselines are publicly available. 
We report the evaluation results in Table~\ref{tab:relevance all} (relevance) and Table~\ref{tab:diversity all} (diversity). As we can observe, on relevance, the pre-trained T5 model that is fine-tuned with the traditional approach (see Section~\ref{sec:transfromers}) generates questions that are more lexical relevant (meaning it generates more relevant question words) compared to previously studied baselines (specifically KPCNet -- the existing state-of-the-art method).  Yet, when it comes to semantic relevance, the same baseline (T5) obtains only competitive question relevance to that of KPCNet. That actually means that, while the model is capable of generating novel (relevant) words (probably due higher vocabulary coverage and better context modeling), the same model does not actually contribute novel questions with new meaning on top of those generated by KPCNet.  This empirical outcome serves as a first evidence to our motivation: traditional fine-tuning of the pre-trained T5 model may result in sub-optimal diversity. 
And indeed, as we can further observe, the T5 baseline has a global diversity that is inferior (in 3 out of the 4 metrics) to that of KPCNet, even though T5 still results in a better local diversity (i.e., lower PW-BLEU[BERTScore] values). 

We next further examine the question generation quality (both with respect to the relevance and diversity metrics) obtained by T5+LTD (i.e., the pre-trained T5 model fine-tuned with our alternative approach). For relevance, we can now observe that, T5+LTD obtains better question relevance (except for the METEOR metric for which it has more or less similar performance to T5), both lexically and semantically. This implies that, fine-tuning the pre-trained T5 model using our LTD approach allows it not only to generate novel words that cover more questions, but also new questions that contribute new meaning to the poll of questions that can be asked over the products collection. This is strongly supported by the diversity metrics obtained by T5+LTD compared to the other baselines, and T5 specifically, where the former significantly outperforms the others in all global diversity metrics (both lexical and semantic). T5+LTD further significantly outperforms all other baselines (accept hMup\footnote{This baseline was mainly designed to obtain high local diversity, yet this comes with the expense of a large relevance drop~\cite{Zhang-KPCNet}.}) on the local diversity metrics. This shows that, while is was mainly designed to improve global diversity (over all products), it is also capable of generating diverse questions on a per-product basis. Overall, these first empirical results demonstrate that, T5+LTD is capable of generating more diverse questions, while preserving the relevance of the underlying T5 Transformer model as much as possible.

\subsubsection{RQ2: Impact of our learning-to-diversify fine-tuning approach}
To answer RQ2, we deepen our comparison between T5 and T5+LTD. We report in Table~\ref{tab:categories} the results of this comparison, when the two alternatives are trained to generate questions for products in different categories. Therefore, such a comparison provides a better analysis of the robustness of our diversification approach considering a variety of product categories. 

We note again that, the LTD approach is applied on the underlying Transformer model \textbf{only during training}, while the inference remains completely \textbf{unchanged}. Hence, in both fine-tuning alternatives, we start from the \textbf{same} pre-trained T5 model and the fine-tuned model is then evaluated on the same grounds. 

As we can observe, for all product categories, applying the LTD approach results in a significant boost in global diversity, both lexically and semantically (up to more than 40\% for some of the metrics). The LTD approach encourages the underlying Transformer model to learn a much richer language model for the task, which results in a more diverse generation. Moreover, in most cases, the local diversity is also improved (specially semantic diversity which always improves). 

Examining the relevance metrics, we can observed that, in the majority of cases, T5+LTD results with a reasonable question generation relevance (for some categories even much better, for some with a relatively slight drop). This serves as another strong empirical evidence that, our LTD approach not only improves the underlying Transformer model diversity (meaning novel questions are being generated), but also preserves, as much as possible, its capability of generating relevant questions. 

\subsubsection{RQ3: Global diversity topic analysis}

\begin{figure}[t!]
    \centering
    \includegraphics[width=0.5\textwidth]{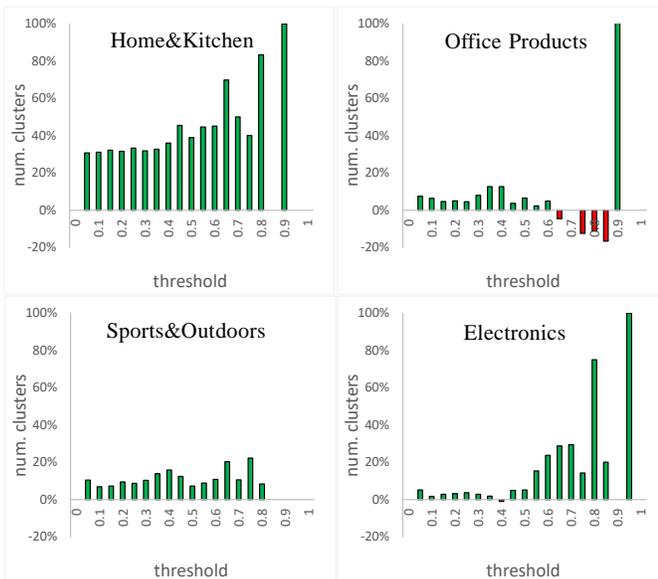}\label{fig:clusters}
\caption{Global diversity topic analysis. The percentages denote the relative improvement (green bars) / degradation (red bars) in the number of clusters per dissimilarity threshold compared to the T5 baseline.}\label{fig:clusters}
\end{figure}

To understand better the impact of our LTD approach on the underlying Transformer model's global diversity, we next perform topic analysis on the generated questions. We hypothesize that, using our LTD approach should result in a significant semantic improvement in terms of number of information needs (``topics'') that can be learned. To this end, using a SentenceTransformer~\cite{reimers2019sentencebert} dedicated for paraphrasing tasks, for each product category, we obtain embeddings for the test-set generated questions of both T5 and T5+LTD. Next, using cosine similarity as the ``distance'' metric, we obtain question clusters\footnote{Clusters obtained with scikit-learn's Agglomerative-Clustering.} for each alternative and measure the number of clusters obtained for increasing (dissimilarity) thresholds. We report the 
results in Figure~\ref{fig:clusters}, illustrating the relative improvement (green bars) or degradation (red bars) of T5+LTD compared to T5. As we can observe, for all product categories, for almost every dissimilarity threshold, T5+LTD obtains significantly more clusters (a statistically validated result $p<10^{-4}$). This serves as another strong empirical evidence for the ability of the LTD approach to enrich the language that can be learned by the underlying Transformer model. 

\subsection{Qualitative Examples}
\begin{table}[t]
\caption{Examples of the top-2 questions generated by T5 and T5+LTD for products in the four categories.}
\center\setlength\tabcolsep{3.0pt}
\small
\begin{tabular}{|l|l|}
\hline
\multicolumn{2}{|l|}{\textbf{Category:} \emph{Home \& Kitchen}}\\
\hline

\textbf{Product} & \multicolumn{1}{l|}{\begin{tabular}[c]{@{}l@{}}\emph{smile rabbit personal ultra-compact air humidifier - cool mist}\end{tabular}}                                                                          \\ \hline

T5               & \multicolumn{1}{l|}{\begin{tabular}[c]{@{}l@{}}does it have a filter ? \\ does this have a filter ?
\end{tabular}}                    
\\ \hline
T5+LTD             & \multicolumn{1}{l|}{\begin{tabular}[c]{@{}l@{}}does this come with a water bottle ? \\ is there a way to turn the mist off ?
\end{tabular}}                                                                           \\ \hline\hline

\textbf{Product} & \multicolumn{1}{l|}{\begin{tabular}[c]{@{}l@{}}\emph{kenmore purple bagless canister vacuum cleaner 22614}\end{tabular}}                                                                          \\ \hline

T5               & \multicolumn{1}{l|}{\begin{tabular}[c]{@{}l@{}}what is the warranty on this vacuum cleaner ? \\ does this vacuum have a warranty ?
\end{tabular}}                  
\\ \hline
T5+LTD             & \multicolumn{1}{l|}{\begin{tabular}[c]{@{}l@{}}what is the warranty on this vac ? \\ does this vacuum have a suction cup ?
\end{tabular}}                                                                                \\ \hline\hline

\multicolumn{2}{|l|}{\textbf{Category:} \emph{Office Products}}\\
\hline
\textbf{Product} & \multicolumn{1}{l|}{\begin{tabular}[c]{@{}l@{}}\emph{quartet whiteboard , white board , dry erase board, 5'x 3',}\\  \emph{silver aluminum frame (s535)}\end{tabular}}                                                                          \\ \hline
T5               & \multicolumn{1}{l|}{\begin{tabular}[c]{@{}l@{}}is this a dry erase board ? \\ is this a dry erase whiteboard ? 
\end{tabular}}     
\\ \hline 
T5+LTD            & \multicolumn{1}{l|}{\begin{tabular}[c]{@{}l@{}}is this whiteboard waterproof ? \\ does it come with a tray ?
\end{tabular}}                                                                               \\ \hline\hline

\textbf{Product} & \multicolumn{1}{l|}{\begin{tabular}[c]{@{}l@{}}\emph{canon imageclass mf3240 monochrome laser all-in-one printer}\end{tabular}}                                                                          \\ \hline

T5               & \multicolumn{1}{l|}{\begin{tabular}[c]{@{}l@{}}does this printer have a usb port ? \\ does it have a usb port ?
\end{tabular}}       
\\ \hline 
T5+LTD             & \multicolumn{1}{l|}{\begin{tabular}[c]{@{}l@{}}does it print both sides of the paper ? \\ does it print in color ?
\end{tabular}}                                                                         \\\hline\hline

\multicolumn{2}{|l|}{\textbf{Category:} \emph{Sports \& Outdoors}}\\
\hline
\textbf{Product} & \multicolumn{1}{l|}{\begin{tabular}[c]{@{}l@{}}\emph{Ozeri 4x3runner Pocket 3D Pedometer and Activity Tracker}\\ \emph{with Dual Walking \& Running Mode Technology}\end{tabular}}                                                                          \\ \hline
T5               & \multicolumn{1}{l|}{\begin{tabular}[c]{@{}l@{}}what is the weight limit for this product ? \\ what is the weight limit for this item ? 
\end{tabular}}  
\\ \hline 
T5+LTD            & \multicolumn{1}{l|}{\begin{tabular}[c]{@{}l@{}}what is the weight limit ? \\ is this waterproof ?
\end{tabular}}                                                                     \\ \hline\hline

\textbf{Product} & \multicolumn{1}{l|}{\begin{tabular}[c]{@{}l@{}}\emph{Powertec Fitness Workbench Utility Bench, Black}\end{tabular}}                                                                          \\ \hline
T5               & \multicolumn{1}{l|}{\begin{tabular}[c]{@{}l@{}}what is the weight capacity of this bench ? \\ what are the dimensions of the bench ?
\end{tabular}}  
\\ \hline 
T5+LTD            & \multicolumn{1}{l|}{\begin{tabular}[c]{@{}l@{}}what is the weight capacity of this bench ? \\ what is the height of the bench when folded ?
\end{tabular}}                                                                   \\ 
\hline\hline

\multicolumn{2}{|l|}{\textbf{Category:} \emph{Electronics}}\\
\hline
\textbf{Product} & \multicolumn{1}{l|}{\begin{tabular}[c]{@{}l@{}}\emph{Dell 2714T 27-Inch Touchscreen LED-lit Monitor}\end{tabular}}                                                                          \\ \hline
T5               & \multicolumn{1}{l|}{\begin{tabular}[c]{@{}l@{}}does this monitor have a backlit keyboard ? \\ does it have a backlit keyboard ? 
\end{tabular}}   
\\ \hline 
T5+LTD            & \multicolumn{1}{l|}{\begin{tabular}[c]{@{}l@{}}does this monitor have a backlit keyboard ? \\ does this monitor have a built in camera ?
\end{tabular}}                                                                   \\ \hline\hline

\textbf{Product} & \multicolumn{1}{l|}{\begin{tabular}[c]{@{}l@{}}\emph{Tabeo E2 8 Inch Kids Tablet - Blue}\end{tabular}}                                                                          \\ \hline
T5               & \multicolumn{1}{l|}{\begin{tabular}[c]{@{}l@{}}does it have a usb port ? \\ does this tablet have a usb port ? 
\end{tabular}}    
\\ \hline 
T5+LTD            & \multicolumn{1}{l|}{\begin{tabular}[c]{@{}l@{}}does it come with a charger ? \\ does it come with a case ? 
\end{tabular}}                                                                   \\ 

\hline

\end{tabular}\label{fig:examples}
\end{table}

We conclude this section with some qualitative examples of questions that were generated by T5 and T5+LTD. Table~\ref{fig:examples} depicts, for each product category, two examples with the top-$2$ questions that were generated by each baseline. Overall, these examples demonstrate two common ``mistakes'' made by the T5 baseline which are mitigated by applying our LTD approach (i.e, T5+LTD). 

First, we note that, although for both fine-tuned model variants we apply the \textbf{same} diversity beam-search method during inference, the questions generated by the T5 baseline still tend to be repetitive (e.g., \emph{does it have a filter ?}, \emph{does this have a filter ?}). While such questions are (more or less) lexically different, they are still semantically equivalent; hence, the second generated question does not actually add any new information. On the other hand, if we examine the questions generated for the same products by T5+LTD, we can observe that its generated questions tend to be less repetitive (if at all). This phenomenon can be attributed to the better local diversity obtained by T5+LTD compared to T5. 

The second notable ``mistake'' made by the T5 baseline is its tendency to generate more general questions (e.g., questions that inquire physical product properties such as dimensions, weight, height, etc, which are usually common to many products in all the four studied categories). On the other hand, examining the questions generated by T5+LTD, we can observe that they are more specific than their counterparts generated by T5. 

Overall, as the qualitative examples demonstrate (and similar to the results we reported in the empirical part), by applying our LTD approach, the pre-trained T5 can be better fine-tuned to enrich its language model for the PQG task, allowing to cover more diverse information needs across many product categories and types.

\section{Summary}\label{sec:summary}
In this work, we have proposed a novel learning-to-diversify (LTD) approach for enhancing the fine-tuning of pre-trained Transformer models for the product question generation task. Using our approach, allows to learn a more globally diverse language model that covers a wider range of user product information needs; this, while preserving the underlying model's generation quality per product in the collection. As future work, we wish to evaluate our approach over other text generation architectures (e.g., Transformer decoder-only models) and tasks.


\balance
\bibliographystyle{ACM-Reference-Format}
\bibliography{qdl-main}

\end{document}